\documentclass[runningheads]{llncs}
\usepackage[T1]{fontenc}
\usepackage{graphicx}
\usepackage{amsmath}
\usepackage{amssymb}
\usepackage{booktabs}
\usepackage{enumitem}
\usepackage{float}
\usepackage{algorithm2e}
\usepackage[dvipsnames,table]{xcolor}
\usepackage{pbox}

\usepackage{xcolor}

\definecolor{ForestGreen}{rgb}{0.13, 0.55, 0.13}
\usepackage{pifont}

\usepackage{enumitem}
\setitemize{noitemsep,topsep=0pt,parsep=0pt,partopsep=0pt,leftmargin=1em}
\setenumerate{noitemsep,topsep=0pt,parsep=0pt,partopsep=0pt,leftmargin=1em}

\DeclareMathOperator*{\argmin}{arg\,min}
\DeclareMathOperator*{\argmax}{arg\,max}
\newtheorem{prop}[theorem]{Proposition}

\newcommand{\trace}{\operatorname{tr}}

\newcommand{\diag}{\operatorname{diag}}

\renewcommand{\paragraph}{\textbf}

\makeatletter
\@tfor\next:=abcdefghijklmnopqrstuvwxyz\do{%
  \def\command@factory#1{%
    \expandafter\def\csname vec#1\endcsname{\mathbf{#1}}
  }
 \expandafter\command@factory\next
}

\@tfor\next:=ABCDEFGHIJKLMNOPQRSTUVWXYZ\do{%
  \def\command@factory#1{%
    \expandafter\def\csname mat#1\endcsname{\mathbf{#1}}
  }
 \expandafter\command@factory\next
}

\@tfor\next:=ABCDEFGHIJKLMNOPQRSTUVWXYZ\do{%
  \def\command@factory#1{%
    \expandafter\def\csname set#1\endcsname{\mathcal{#1}}
  }
 \expandafter\command@factory\next
}

\def\greekvectors#1{%
 \@for\next:=#1\do{%
    \def\X##1;{%
     \expandafter\def\csname mat##1\endcsname{\boldsymbol{\csname##1\endcsname}}
     }
   \expandafter\X\next;
  }
}

\greekvectors{alpha,beta,iota,gamma,lambda,nu,eta,Gamma,varsigma,Phi,Theta}

\makeatother

\begin{document}
\title{Non-Negative Spherical Relaxations for Universe-Free Multi-Matching and Clustering
}
\titlerunning{Non-Negative Spherical Multi-Matching and Clustering}
\author{
Johan Thunberg\inst{1}
\and
Florian Bernard\inst{2}
}
\authorrunning{}
\institute{
Halmstad University, Sweden\\
\and
University of Bonn, Germany
}

\maketitle              

\begin{abstract}
We propose a novel non-negative spherical relaxation for optimization problems over binary matrices with injectivity constraints, which in particular has applications in multi-matching and clustering. We relax respective binary matrix constraints to the (high-dimensional) non-negative sphere. To optimize our relaxed problem, we use  a conditional power iteration method to iteratively improve the objective function, while at same time sweeping over a continuous scalar parameter that is (indirectly) related to the universe size (or number of clusters). Opposed to existing procedures that require to fix the integer universe size before optimization, our method automatically adjusts the analogous continuous parameter. Furthermore, while our approach shares  similarities with spectral multi-matching and spectral clustering, our formulation has the strong advantage that we do not rely on additional  post-processing procedures to obtain binary results. Our method shows compelling results in various multi-matching and clustering settings, even when compared to methods that use the ground truth universe size (or number of clusters).

\end{abstract}

\section{Introduction}
In this work we  propose a novel 
method that generalizes the power iteration algorithm to handle non-negative matrices with rows on the unit sphere (i.e. the intersection of the non-negative orthant and the unit sphere), referred to as ``non-negative sphere''. This is highly relevant for optimization problems that aim to find assignments between elements, such as they occur in multi-matching (e.g.~for finding correspondences between keypoints in a collection of images~\cite{Pachauri:2013wx}), or in clustering problems, where points are to be grouped according to some similarity criterion~\cite{jain2010data}. While there is a range of approaches based on spectral matrix decompositions, such as spectral permutation synchronization~\cite{Pachauri:2013wx,Shen:2016wx,Maset:YO8y6VRb,bernard2021sparse} or spectral clustering~\cite{von2007tutorial}, existing procedures have the downside that they require that the universe size (or number of clusters) $k$ is known, and additionally require post-processing to find binary solutions representing discrete assignments.

Our method circumvents these shortcomings by using a similarity matrix to group vectors on the $(k{-}1)$-dimensional non-negative sphere, where $k$ is an (arbitrarily loose) upper bound for the universe size (i.e.~the total number of unique elements across all objects in multi-matching problems, or the number of clusters in clustering problems). Our procedure shares resemblance with power iterations and orthogonal iterations for eigenspace computation~\cite{golub2013matrix}, but optimizes over 
non-negative  matrix \emph{rows} with unit length (instead of orthogonal \emph{columns}).
In addition, we automatically determine a single real-valued parameter that is (indirectly) related to the (unknown) integer number $k$, 
which we address by utilizing specific properties that directly arise from our novel formulation.
We summarize our main contributions as follows: 
\begin{itemize}
    \item Our formulation does not require the  knowledge of $k$. Instead,
    we simply require to select an (arbitrarily loose) upper bound of $k$, which corresponds to the number of columns of our matrix optimization variable. We emphasize that it is not possible to choose a too large upper bound that would impair the result quality.
    
    \item Opposed to related existing methods, our method does not require any additional post-processing in order to find discrete assignments. 
    \item
    We show that our procedure
    leads to state-of-the-art results in diverse  multi-matching and clustering problems.
\end{itemize}

\section{Related Work}
In the following we provide an overview of the relevance of spectral methods for multi-matching and clustering.

\textbf{Multi-matching and permutation synchronization.} Multi-matching refers to the problem of finding correspondences across a collection of objects (e.g.~correspondences between keypoints in images, or between vertices of meshes, etc.). Many approaches are based on permutation synchronization~\cite{Pachauri:2013wx,Shen:2016wx,Maset:YO8y6VRb,huang2019learning,birdal2019probabilistic,bernard2019synchronisation,huang2019tensor,birdal2021quantum,bernard2021sparse}, which is a procedure that establishes cycle consistency in the set of pairwise matchings. The main idea is to represent pairwise matchings in a large block matrix, and then use low-rank matrix factorization (e.g.~via spectral decompositions, convex or non-convex optimization) to extract cycle-consistent matchings. Similarly, analogous strategies have been used for other types of pairwise transformations, such as functional maps~\cite{huang2020consistent,gao2021isometric}, or spatial transformations~\cite{bernard2015solution,arrigoni2016spectral}. While permutation synchronization can be interpreted as a multi-object version of the linear assignment problem~\cite{Munkres:1957ju}, respective approaches generally do not consider pairwise terms and thereby are unable to take into account geometric consistency. Instead, to consider geometric consistency, a multi-object version of the quadratic assignment problem~\cite{Koopmans:1957gf,Lawler:1963wn} can be considered, which, however, leads to an objective function that is a fourth-order polynomial. The latter was successfully addressed based on a higher-order projected power iteration method~\cite{bernard2019hippi}. All of the mentioned approaches require that the universe size
is known (or accurately estimated). In contrast, in this work we propose an approach that does not rely on this information.

\textbf{Spectral clustering and related work.} Clustering is  well studied and thousands of algorithms have appeared~\cite{jain2010data} since the k-means algorithm was introduced over sixty years ago~\cite{steinhaus1956division}.
Spectral clustering was first developed in the context of graph partitioning~\cite{fiedler1973algebraic}, and a large number of related approaches have been  proposed~\cite{shi2000normalized,donath2003lower,kannan2004clusterings,ng2002spectral,ding2005equivalence,bach2006learning,zhang2008multiway,Lu:2014ht}.
Essentially, in spectral clustering the intractable NP-hard clustering problem~\cite{drineas2004clustering} is ``relaxed'' to a tractable eigenvector problem, whose solution is then eventually rounded to approximate a solution of the original problem. 
Typically, pairwise distances between data points are computed first, which are then represented as a distance matrix, i.e.~an object that reflects dissimilarities between data points. Subsequently, a kernel function is applied to the distance matrix~\cite{zelnik2004self,zhang2011local,wang2014similarity,wang2017visualization} in order to convert it  to a similarity matrix. Kernel functions range from fixed Gaussians~\cite{ng2002spectral} to adaptive approaches~\cite{wang2017visualization,john2020spectrum}. Once a suitable similarity matrix is obtained, eigenvectors are computed as a low-dimensional embedding of the similarity matrix, which is then used for clustering. 
A common way to compute spectral clustering solutions is via power iterations or orthogonal iterations~\cite{golub2013matrix}. 
Spectral formulations furthermore appear as relaxations to normalized cut and min-cut formulations for clustering~\cite{chan1994spectral,stella2003multiclass}. There is also a connection to kernel-based k-means~\cite{dhillon2004kernel}. Recently there have been several works that incorporate learning of balance parameters in the spectral clustering formulation~\cite{chen2020enhanced,chen2017self,shaham2018spectralnet}.

\textbf{Spherical clustering} refers to the clustering of points distributed over the (unit) sphere.
Two common methods for spherical clustering are spherical k-means~\cite{dhillon2001concept} and online spherical k-means~\cite{zhong2005efficient}. The k-means clustering method~\cite{steinhaus1956division,duda1973pattern} is a special case of a Gaussian mixture model in which all priors are equal for the components, and all parameters are equal for the concentrations.
The spherical k-means methods share analogous properties with respect to mixture models based on the von Mises-Fisher distribution on the unit sphere~\cite{banerjee2002frequency,banerjee2003clustering}, which uses cosine-functions of angular differences. 
This transitions into generative model-based (or parametric models) for spherical clustering, for which many approaches have been suggested~\cite{zhong2003comparative,dortet2008model,kent1982fisher,bijral2007mixture,sra2013multivariate}. In this context the Expectation Maximization algorithm is commonly used~\cite{banerjee2003clustering,golzy2016algorithms}. While spherical clustering approaches have the objective to cluster points distributed over the sphere, in our approach we consider the clustering of points in any space. To this end, we consider a conditional power iteration method~\cite{frank1956algorithm,jaggi2013revisiting} that iteratively finds points on the non-negative sphere in order to separate them into clusters.

\textbf{Generalizations of power iterations.} 
Due to its simplicity and its high practical value, there have been a diverse range of generalizations of the power iteration method. Among them are the higher-order power method~\cite{de1995higher},
the incorporation of additional penalties~\cite{journee2010generalized}, a  variant for nonlinear eigenproblems~\cite{hein2010inverse},
a max-pooling-based variant~\cite{cho2014finding}, tensor power iterations~\cite{shi2016tensor}, coordinate-wise power iterations~\cite{lei2016coordinate}, projected power iterations~\cite{chen2018projected},  higher-order projected power iterations~\cite{bernard2019hippi}, or a sparsity-inducing formulation~\cite{bernard2021sparse}.
Moreover, the power iteration algorithm can also be interpreted as a special case of the Frank-Wolfe algorithm~\cite{frank1956algorithm,jaggi2013revisiting} applied to Rayleigh quotient optimization.

\section{Preliminaries}  
We define the set of binary row-stochastic matrices with $m$ rows and $k$ columns as $\mathcal{P}_{m,k}$. Another description of $\mathcal{P}_{m,k}$ is that it comprises $m$-by-$k$ matrices with rows equal to the canonical  basis vectors in $\mathbb{R}^k$.
Suppose a data matrix $X \in \mathbb{R}^{m \times k}$ is given, where the rows represent a total of $m$ data points in $k$ dimensions. A similarity matrix $W \in \mathbb{R}^{m\times m}$ for $X$ is a non-negative symmetric matrix, where each element $W_{ij}$ captures the similarity (in some appropriate sense) between the row $i$ (i.e. data point $i$) and row $j$ (i.e. data point $j$) of $X$. Larger values indicate higher similarity, and smaller values indicate lower similarity.

Consider the nonlinear least-squares optimization problem on the form 
\begin{equation}
\label{eq:X:1}
   \argmin_{U \in \mathcal{P} \subset \mathcal{P}_{m,k}} f(\beta, W,U), 
\end{equation}
where $\mathcal{P}$ is a subset of $\mathcal{P}_{m,k}$ that depends on the type of problem we want to solve, and  
\begin{align}
    \label{eq:new:1}
     f(\beta, W, U) = \|W {-} \beta UU^T\|_{\text{F}}^2 
    = \trace(W^2) {-} 2\beta\trace(U^TWU) {+} \beta^2\trace(UU^TUU^T),
\end{align}
where 
$\beta > 0$ is some parameter of choice. Note that this formulation only involves $W$ (and not a data matrix $X$), which could have been obtained without using a data matrix $X$ in applications such as graph clustering or permutation synchronization. 

As mentioned, the choice of $\mathcal{P}$ in \eqref{eq:X:1} depends on the type of problem considered. In clustering, $\mathcal{P}$ is equal to $\mathcal{P}_{m,k}$. In multi-matching, $\mathcal{P}$ has additional structure. In such cases $\mathcal{P}$ is the set of matrices on the form $U = [U_1^T, U_2^T, \ldots, U_q]^T$, where $q < m$ and each $U_i$ is a partial permutation matrix, i.e.~$U_i \in \mathbb{P}_{m_iq}$ with
\begin{equation}
    \mathbb{P}_{m_iq} := \{X \in \{0, 1\}^{m_i \times k } : X\bold{1}_{k} \leq  \bold{1}_{m_i}, \:  \bold{1}_{m_i}^TX \leq  \bold{1}_{k}^T\} \quad \text{and} \quad \sum_{i= 1}^qm_i = m.
\end{equation}

\subsection{Non-Negative Spherical Relaxation}\label{sec:nnsr}
\begin{figure}[h!]
\center
  \includegraphics[scale=0.50]{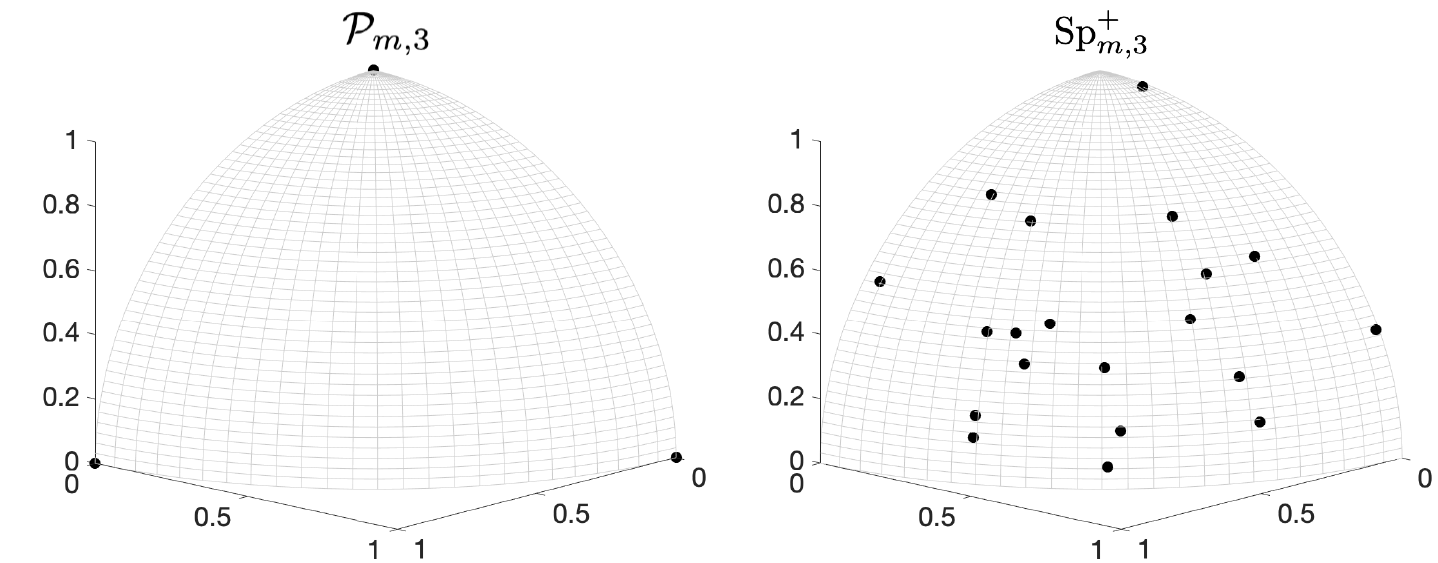}
  \caption{\textbf{Left:} points in the set $\mathcal{P}_{m,3}$ are corner points on the non-negative sphere. \textbf{Right:} points in the relaxed set $\text{Sp}^+_{m,3}$ lie on the non-negative sphere.} 
  \label{fig:sphereRelax}
\end{figure}

We revisit equation \eqref{eq:new:1} and note that for $U \in \mathcal{P}_{m,k}$ it holds that 
\begin{align}
    \trace(UU^TUU^T) = \trace((U^TU)(U^TU)) 
    =  \trace(U^T\bold{1}_{m,m}U),
\end{align}
where 
$\bold{1}_{m,m}$ is the matrix in $\mathbb{R}^{m \times m}$ whose elements are all $1$. 
This means that minimization of $f$ in \eqref{eq:new:1} over $\mathcal{P}_{m,k}$ may equivalently be expressed as maximization of 
\begin{equation}
\label{eq:new:2}
    2\beta\trace(U^TWU) - \beta^2\trace(U^T\bold{1}_{m,m}U) 
\end{equation}
over $\mathcal{P}_{m,k}$, since the constant term in \eqref{eq:new:1} can be ignored. We introduce the quadratic form formulation as
\begin{equation}
\label{eq:new:3}
    \argmax_{U \in \mathcal{P}_{m,k}} g(\alpha, W,U),
\end{equation}
where 
\begin{equation}
    g(\alpha, W,U) = (1-\alpha)\trace(U^TWU) - \alpha \trace(U^T\bold{1}_{m,m}U).
\end{equation}
We see that \eqref{eq:new:3} is equivalent to maximization of  \eqref{eq:new:2} by setting $\alpha =  \beta^2/(\beta^2 + 2\beta) > 0$. The  maximization of \eqref{eq:new:2} is in turn equivalent to minimization of \eqref{eq:new:1}, as stated above. 
Now we define its \textbf{non-negative spherical relaxation} as

\begin{equation}
\label{eq:new:4}
    \argmax_{U \in \text{Sp}^+_{m,k}} g(\alpha, W,U),
\end{equation}
where $\text{Sp}^+_{m,k} = \{[u_1^T, u_2^T, \ldots, u_m^T]^T \in \mathbb{R}^{m \times k}: u_i \geq 0 \text{ and }  u_iu_i^T = 1 \: \forall \: i\} $, i.e. each row $u_i$ of a matrix $U$ in $\text{Sp}^+_{m,k}$ has non-negative elements and unit length (a point on the non-negative part of the unit sphere).  
An illustration of this relaxation is provided in Fig.~\ref{fig:sphereRelax} for $m=3$. 

\subsection{Conditional Power Iteration}\label{sec:power}
We define two operations for re-balancing the similarity matrix $W$ before used in the optimization problem, the \textbf{normalization} 
\begin{equation}
\label{eq:new:5}
    W \leftarrow DWD, \text{ where } D = \sqrt{m}\diag(W\bold{1}_{m})^{-\frac{1}{2}};
\end{equation}
and the 
\textbf{$\kappa$-shift} 
\begin{equation}
\label{eq:new:111}
     W \leftarrow W + \kappa \bold{1}_{m,m}, 
\end{equation}
for a $\kappa > 0.$
There is an efficient  (local) optimization method for \eqref{eq:new:4} (with the re-balanced $W$) referred to as \emph{conditional power iterations}, see Algorithm~\ref{alg:1}. In the algorithm we use the matrix 
\begin{equation}
    V(\alpha) = \delta(\alpha) \bold{I}_m +  (1-\alpha) W - \alpha\boldsymbol{1}_{m,m},
\end{equation}
where $\delta(\alpha) \geq 0$ is some function of $\alpha$ such that $V(\alpha)$ is positive definite. We will throughout assume $\delta$ is given by minus the smallest eigenvalue of $(1-\alpha)W - \alpha\boldsymbol{1}_{m,m}$ plus some additional positive epsilon.

\begin{algorithm}
\SetKwInput{Input}{Input}
\SetKwInput{Output}{Output}
\SetKwInput{Initialise}{Initialise}
\SetKwRepeat{Do}{do}{while}
\DontPrintSemicolon
 \Input{$U_0 \in \mathcal{S}$}
 \Output{$U$}
\Initialise{$t \gets 0$}
  \Repeat{ convergence}{ 
   $U_{t+1} \gets \argmax\limits_{\tilde{U} \in \mathcal{S}}~ \text{tr} (U_t^T V(\alpha) \tilde{U} )$ \\
    $t \gets t{+}1$
   }
   $U \gets U_t$
   
 \caption{Conditional power iterations}\label{alg:1}
\end{algorithm}

The algorithm resembles the conditional gradient method (also known as the Frank-Wolfe method~\cite{jaggi2013revisiting}), for objectives of quadratic form, where we allow for $\mathcal{S}$ to be any (not necessarily convex) compact subset of $\mathbb{R}^{m \times k}$. 
In the following, we summarize convergence properties of Algorithm~\eqref{alg:1}.
\begin{prop}\label{lem:1}
The following holds for the conditional power iterations in Algorithm~\eqref{alg:1}. 
\begin{enumerate}
\item $\text{tr}(U_{t+1}^TV(\alpha)U_{t+1}) \geq \text{tr}(U_{t}^TV(\alpha)U_{t})$. 
\item There is $\delta > 0$, s.t. $\text{tr}(U_{t}^TV(\alpha)U_{t}) \rightarrow \delta$ as $t \rightarrow \infty$, i.e. the objective converges. 
\item For any $\epsilon > 0$ there is a $T \in \mathbb{N}$ (as a function of $\epsilon$) s.t.~$\sum_{t = T}^{\infty}\|U_{t+1} - U_t\|^2_F \leq \epsilon$.
\end{enumerate}
\end{prop}

A proof of Proposition~\ref{lem:1} is provided in the appendix. For manifolds one can, in general, further strengthen condition 3:  the solution converges to a critical point, and converges almost everywhere to a local maximum \cite{lee2019first}. \\

\noindent
\paragraph{Using Algorithm~\ref{alg:1}:} when $\mathcal{S} = \text{Sp}^+_{m,k}$, we immediately note that the maximizers of \eqref{eq:new:4} are equal to those of $\text{tr} (U^T V(\alpha) {U} )$. This can be observed by noting that
    $\text{tr} (U^T(\delta(\alpha) \bold{I}_m) {U} ) = \delta(\alpha)\text{tr}(UU^T)  
     = \delta(\alpha)\text{tr}(U^TU)= \delta(\alpha)m$ is constant for $U \in \text{Sp}^+_{m,k}$.

The main update in Algorithm~\ref{alg:1} reads
\begin{align}\label{eq:update}
    U_{t+1} \gets  \argmax\limits_{\tilde{U} \in \text{Sp}^+_{m,k}} \text{tr} (U_t^T V(\alpha) \tilde{U} ).
\end{align}
Since the rows of matrices in the set $\text{Sp}^+_{m,k}$ are independent,  the maximization in~\eqref{eq:update} can be solved by performing maximizing for each column of $U_t^TV$ individually. An efficient solution to this per-column maximization problem is presented in Algorithm~\ref{alg:3}.

\begin{algorithm}
\SetKwInput{Input}{Input}
\SetKwInput{Output}{Output}
\SetKwInput{Initialise}{Initialise}
\SetKwRepeat{Do}{do}{while}
\DontPrintSemicolon
 \Input{$x \in \mathbb{R}^k$}
 \Output{$y \in \text{Sp}^+_{1,k}$}
  \eIf{$\max_i x_i \leq 0$ }
  {
    $y \gets \boldsymbol{0}_k$\\
    $i = \argmax_i x_i$ \tcp{select any solution if not unique}
    $y_i \gets 1$ 
  }
  {
   $y \gets \max(x, 0) \text{ (element-wise maximization) }$ \\
  $y  \gets \frac{y}{\|y\|_2} $
  }
 \caption{Projection onto non-negative sphere}\label{alg:3}
\end{algorithm}

In the following we assume that $\mathcal{S} = \text{Sp}^+_{m,k}$ and that the maximization problem in \eqref{eq:update} is solved using Algorithm~\ref{alg:3}.
Given the output $U$ to Algorithm~\ref{alg:1}, we subsequently perform a rounding to obtain a matrix $\tilde U \in \text{Sp}^+_{m,k} \cap \mathcal{P}_{m,k}$, from  $U \in \text{Sp}^+_{m,k}$. 

\section{Universe-Size-Free Procedure}
In Sec.~\ref{sec:power} we provided a strategy for local optimization of \eqref{eq:new:4} for a given $\alpha$. The problem is that this parameter $\alpha$ is not given. It does not make sense to, for example, choose  $\alpha=1$, since then the optimal solution $U^*$ does not depend on $W$. 
The main approach in this section is to select a ``good'' $\alpha$ by using the convergence rate of Algorithm~\ref{alg:1} for different choices of $\alpha$. Regarding the choice of $k$, our only requirement in the proposed procedure is that it is sufficiently large.

\subsection{Description of the method}
\begin{figure}[h!]
\center
  \includegraphics[scale=0.48]{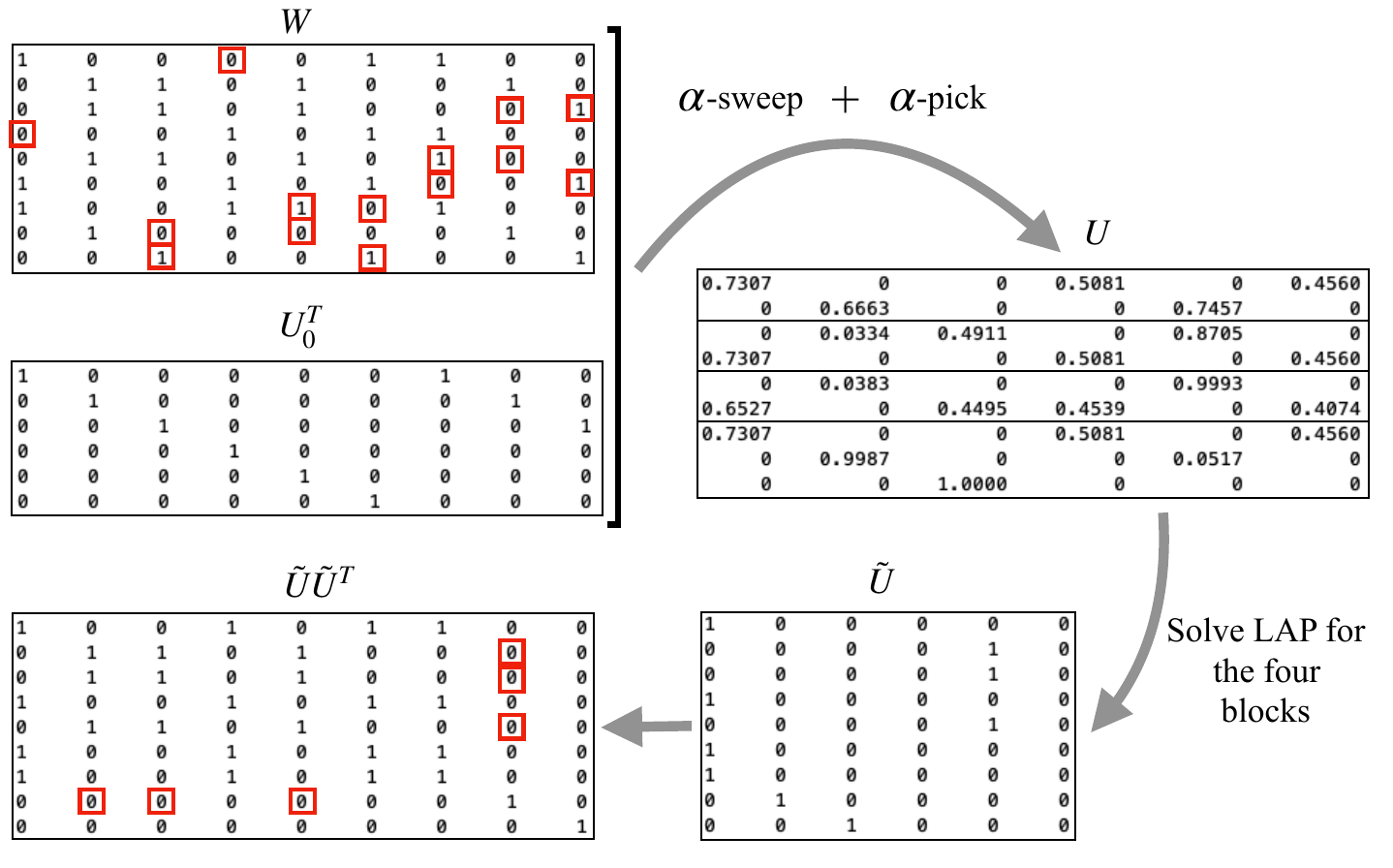}
  \caption{Illustration of the method for a small multi-matching problem. Universe size (number of unique points) is $k^* = 3$, number of images is 4, each containing either 2 or 3 point points in correspondence with points in other images. The sum of all points (in all objects) is $m = 9$. We choose $k = 6$ for the method. There are two inputs to the method: the noisy similarity matrix $W$, where erroneous entries are marked by red; and the matrix $U_0$.  The main procedure $\alpha$-sweep 
 (Algorithm~\ref{alg:4}) and $\alpha$-pick 
 (Algorithm~\ref{alg:5}) produces 
 a matrix $U \in \text{Sp}^+_{9,6}$. The linear assignment problem (LAP) is solved for each of the four blocks of $U$ to produce the matrix $\tilde{U}$, which defines the correspondences. 
 } 
  \label{fig:illustration11}
  \vspace{-3mm}
\end{figure}
\begin{figure}[!]
\center
  \includegraphics[scale=0.25]{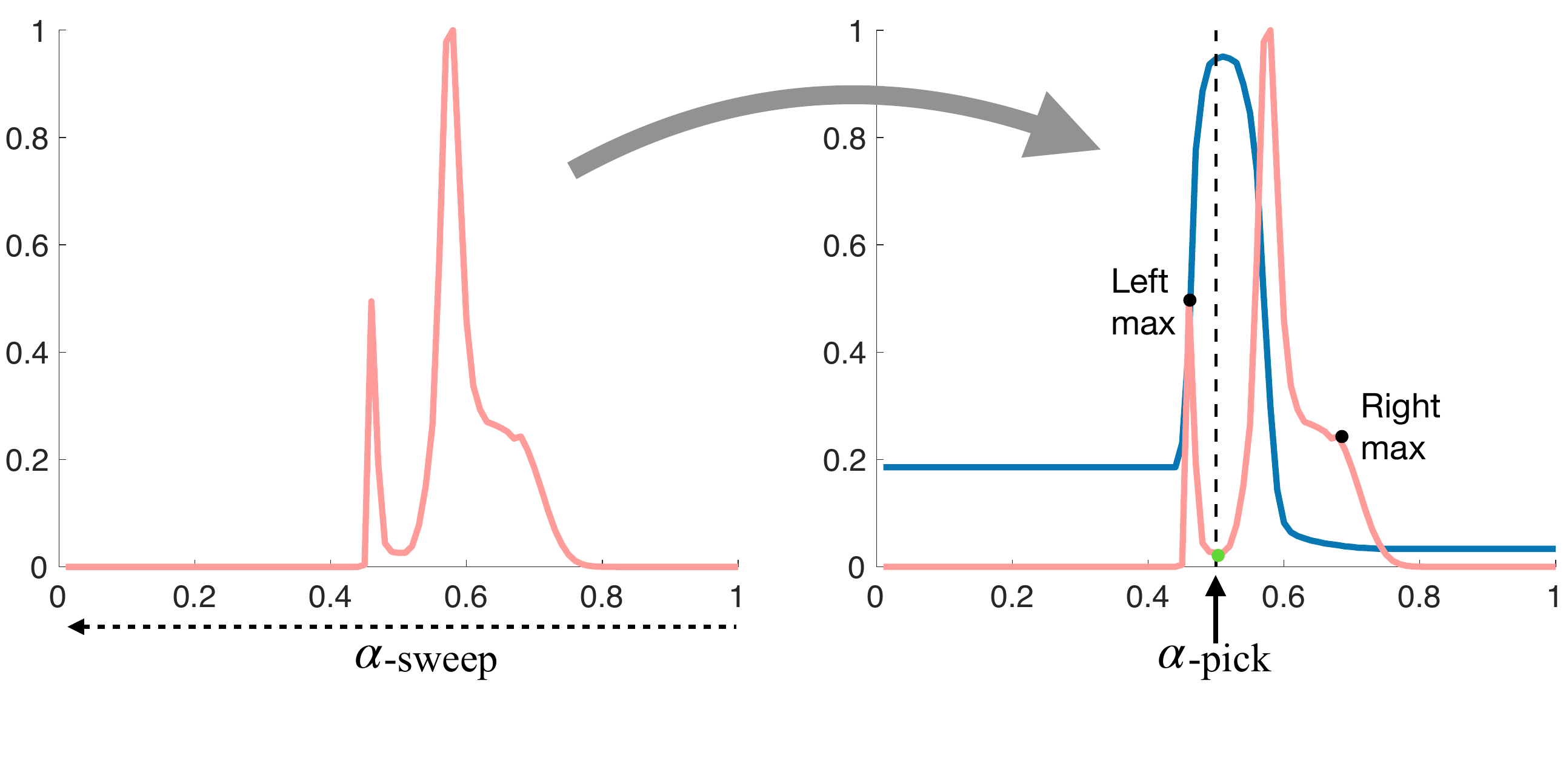}
  \vspace{-8mm}
  \caption{The method is used for a sample clustering problem for illustration. \textbf{Left}: the computed measure of convergence rate $\eta$ (pink) as a function of $\alpha$ for  Algorithm~\ref{alg:4} (i.e. $\alpha$-sweep). \textbf{Right}: an $\alpha$-value is picked for the (pink) $\eta$-curve using Algorithm~\ref{alg:5} (i.e. $\alpha$-pick). For illustration, the F-score for the solution for each $\alpha$ is shown in blue and the dashed line shows the F-score value of the solution corresponding to the picked $\alpha$. }
  \label{fig:illustration}
\end{figure}

\noindent
 First the matrix $W$ is normalized, see \eqref{eq:new:5}, and consecutively $\kappa$-shifted, see \eqref{eq:new:111}, (with some small $\kappa > 0$). The procedure is then to run two consecutive algorithms: $\alpha$-sweep, see Algorithm~\ref{alg:4} below, and $\alpha$-pick, see Algorithm~\ref{alg:5} below. The method is illustrated on a high level for a multi-matching problem in Fig.~\ref{fig:illustration11}. The main intuition behind the algorithms is illustrated for a clustering problem in Fig.~\ref{fig:illustration}. \\

 \noindent
 \paragraph{$\alpha$-sweep:} Initially we choose $\alpha = 1$ and $k$ to some sufficiently large  positive integer. We choose $U_0$ to the unique matrix $K \in \mathbb{R}^{m, k}$ for which there is a matrix $R$ and integer $\bar m > m$ such that $[K^T, R^T]^T = [I_k, I_k, \ldots, I_k]^T \in \mathbb{R}^{\bar m \times k}$. In other words, $U_0$ comprises the upper part (the first $m$ rows) of a tall matrix with repeated identity matrix blocks. 
 Then we decrease $\alpha$ (from $1$) to $0$ with a specified step-size $\epsilon_{\alpha}$. For each $\alpha$ considered on the path from $1$ to $0$, we run the conditional power iteration Algorithm~\ref{alg:1} (outlined in Section~\ref{sec:power}) for a few fixed number of steps to update $U$. Thus, when we update $\alpha$ with the step size $\epsilon_{\alpha}$, we use the final $U$ obtained from for the previous $\alpha$ as initialization. 
 For each $\alpha$ considered, we compute the difference between the objective function with $U$ from the last iteration and the objective function with $U$ from the second last iteration. We call this difference $\eta(\alpha)$ and it is a measure of how fast the objective converges for each considered $\alpha$. \\

\begin{algorithm}[h!]
\SetKwInput{Input}{Input}
\SetKwInput{Output}{Output}
\SetKwInput{Initialise}{Initialise}
\SetKwRepeat{Do}{do}{while}
\DontPrintSemicolon
 \Input{${W}$, $k$, $\epsilon_{\alpha}$, $\epsilon_{\eta}$ $N$ $k$}
 \Output{$\alpha_{tot}$, $\eta_{tot}$, $U_{tot}$}
\Initialise{$\alpha \gets 1$, $\alpha_{tot} \gets [\:]$, $h_{tot} \gets [\:]$, $U_0 \gets K$, $U_{tot} \gets U_0$}
\tcp{outer-loop sweeps discretely over $\alpha$} 
  \Repeat{$\alpha - \epsilon_{\alpha} < 0$}{ 
   $\alpha \gets \alpha - \epsilon_{\alpha}$ \\
   $\lambda_{\min} \gets \min(\text{eig}((1-\alpha) W - \alpha\boldsymbol{1}_{m,m}))$ ~\tcp{ smallest eigenvalue}
   $V \gets (\lambda_{\min} + \epsilon_{\eta}) \bold{I}_m +  (1-\alpha) W - \alpha\boldsymbol{1}_{m,m}$
  
   $i \gets 0$, \\
   \tcp{inner-loop uses Algorithm 1}
    \Repeat{$i = N$}{ 
    $U_{t+1} \gets  \argmax\limits_{\tilde{U} \in \text{Sp}^+_{m,k}}~ \text{tr} ( U_t^T V(\alpha) \tilde{U} )$  \\
    $i \gets i + 1$ \\
    $t \gets t + 1$
    }
     $\alpha_{\text{tot}} \gets [\alpha_{\text{tot}}, \alpha]$ \\
     $U_{tot} \gets [U_{tot}|U_{t}]$ \\
   $\eta_{\text{tot}} \gets [\eta_{\text{tot}}, |g(\alpha,  W, U_{t}) - g(\alpha,  W, U_{t-1})|]$ \\
   }
   
 \caption{$\alpha$-sweep}\label{alg:4}
 
\end{algorithm}
 
 \noindent
 \paragraph{$\alpha$-pick:} This proceedure is illustrated in the right sub-figure of Fig.~\ref{fig:illustration}. We find the first local maximum of $\eta$ from the right, referred to as ``Right max'' in the figure. Then we find the first local maximum of $\eta$ from the left, referred to as ``Left max'' in the figure. Then we pick $\alpha^*$ as the corresponding $\alpha$ for the minimum $\eta$ between ``Left max'' and ``Right max'', shown with a green dot in the figure. In the figure we also show the F-score (blue line), which is the harmonic mean of precision and recall. The black dashed line illustrates the corresponding F-score for the picked $\alpha^*$. The procedure for finding the index of the value $\alpha^*$ in the $\alpha$-vector is formally described in Algorithm~\ref{alg:5}. \\

\begin{algorithm}[!]
\SetKwInput{Input}{Input}
\SetKwInput{Output}{Output}
\SetKwInput{Initialise}{Initialise}
\SetKwRepeat{Do}{do}{while}
\DontPrintSemicolon
 \Input{$\alpha_{\text{tot}}$ \tcp{from  Algorithm 4}}
 \Output{$\text{i}^*$}
\Initialise{$i_r \gets 0$, $i_l \gets \text{length}(\eta_{\text{tot}})$~\tcp{length of the vector}}
  \Repeat{$\eta_{\text{tot}}(i_r + 1) < \eta_{\text{tot}}(i_r)$ ~\tcp{value at subsequent index in vector is smaller}} 
  { 
    $i_r \gets i_r + 1$ 
   }

  \Repeat{$\eta_{\text{tot}}(i_l - 1) < \eta_{\text{tot}}(i_l)$ ~\tcp{value at previous index in vector is smaller}}{ 
   $i_l \gets i_l - 1$ 
   }
   $\text{i}^* \gets i_r + \text{minIndex}(\eta_{\text{tot}}(i_r:i_l))$ ~\tcp{ index of minimum value between the two maxima} 
 \caption{$\alpha$-pick}\label{alg:5}
 
\end{algorithm}

 \subsection{Brief Motivation}
We begin by addressing initialization of $U_0$ and why $\alpha$ is decreased from $1$ to $0$ in the $\alpha$-sweep algorithm. For $\alpha = 1$, one can show that  
$K$ is an optimal solution to \eqref{eq:new:3}. 
Optimal solutions for $\alpha = 0$ comprise matrices with all rows equal. But, the optimal solutions for $\alpha = 0$ comprise a critical set for all $\alpha \in [0,1]$. Hence the solution will not change when $\alpha$ is increased from $0$ to $1$, as we would always get the same trivial solutions when varying $\alpha$ from $0$ to $1$. 
The pink curve in Fig.~\ref{fig:illustration} would be replaced by a straight line. 
Now, with a $\kappa$-shift, the optimal solutions for $\alpha > 0$ small enough are the matrices with all rows equal. The objective will be the same when varying $\alpha$ in that region. Thus the $\eta$-values will be close to zero in that region, see Fig.~\ref{fig:illustration} when $\alpha$ is smaller than $0.43$. If the similarity  
matrix is ``free of noise'', there is a region of $\alpha$-values for which the optimal solution is retrieved and $\eta$ on that interval should be close to $0$ (convergence of objective is assured by Proposition~\ref{lem:1}). The objective value, however, is different from that when $\alpha = 0$ or $\alpha = 1$. By varying $\alpha$ between $1$ and $0$, one should first experience an increase in $\eta$, then a local minimum, then an increase, and then a local minimum again.

\section{Experiments}\label{sec:experiments}

\subsection{Multi-Image Matching}
Multi-image matching refers to the problem of establishing correspondences between given keypoints in a collection of images~\cite{Pachauri:2013wx,zhou2015multi,yan2016short,tron2017fast,swoboda2019convex,bernard2019hippi,birdal2019probabilistic,birdal2021quantum}. 
 Our procedure for these type of problems is illustrated with an example in Fig.~\ref{fig:illustration11}. As rounding, we solve $q$ (partial) linear assignment problems for the $q$ blocks in the output matrix $U$. \\

\begin{figure*}[!h]
\centering
  \includegraphics[scale=0.44]{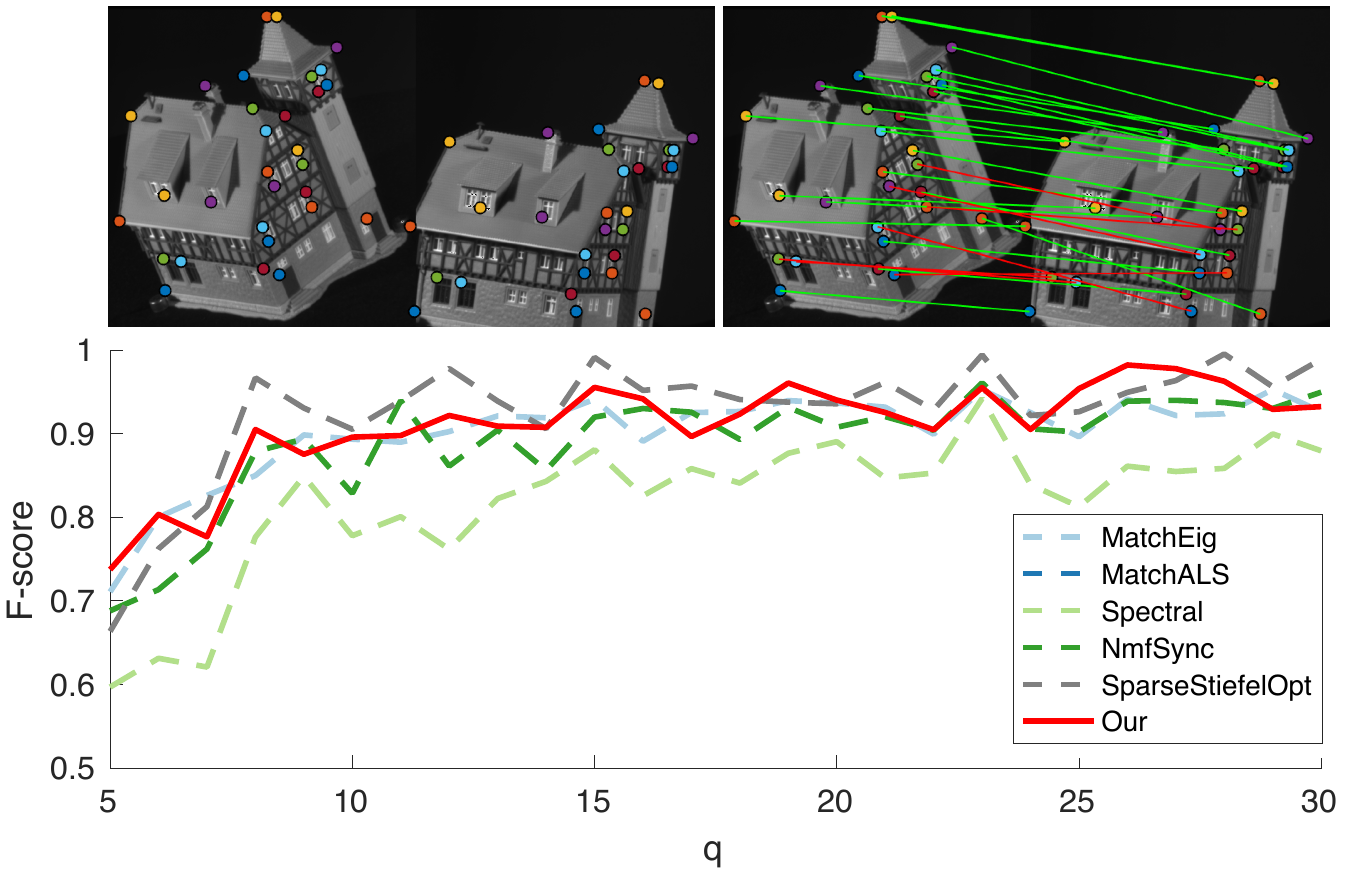}~~
  \vspace{-5mm}
  \caption{\textbf{Top left}: two images from the CMU house image sequence. \textbf{Top right}: example of correct vs. incorrect pairwise matchings. Green lines show correct matchings, whereas red lines show incorrect matchings. \textbf{Bottom}: quantitative comparison of state-of-the-art multi-image matching methods in terms of the F-Score. All methods with dashed lines \emph{use the ground-truth universe size for matching}, whereas \textbf{Our}  (solid) does not.
 }
  \label{fig:x1}
\end{figure*}

\noindent
\paragraph{CMU house image sequence}.
We begin by considering the CMU house image sequence~\cite{cmuHouse}, from which we generate a range of multi-image matching problems following the procedure of~\cite{Pachauri:2013wx}. We compare our method with \textbf{MatchEig}~\cite{Maset:YO8y6VRb}, \textbf{Spectral}~\cite{Pachauri:2013wx},  \textbf{SparseStiefelOpt}~\cite{bernard2021sparse},  \textbf{MatchALS}~\cite{zhou2015multi}, \textbf{NmfSync}~\cite{bernard2019synchronisation}.
In this experiment for \textbf{Our} 
we use $k=100$, $\epsilon_{\alpha} = 0.01$, $N=5$, $\epsilon_{\eta} = 0$, $\kappa = 0.01$. 
We present results in Fig.~\ref{fig:x1} in terms of F-score, i.e. harmonic mean of precision and recall. \\

\begin{figure*}[!h]
  \includegraphics[width=0.95\linewidth]{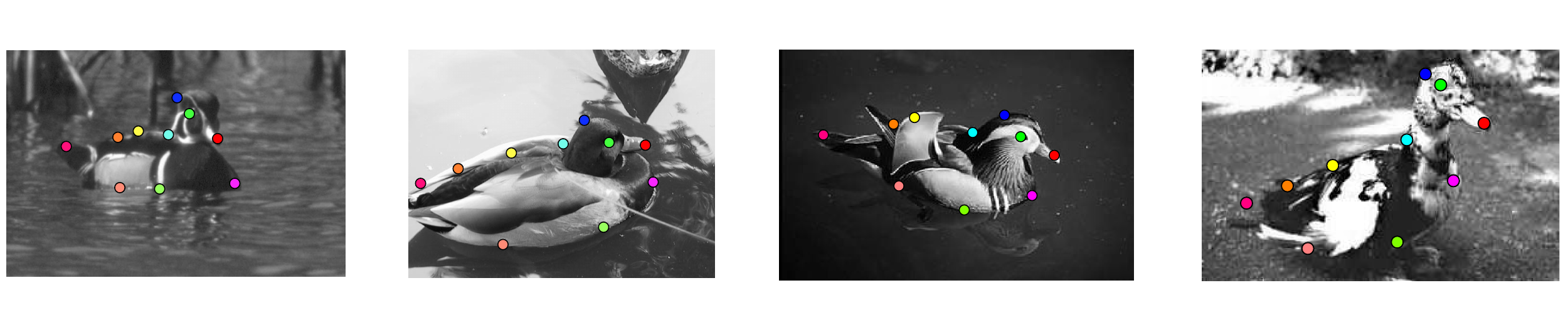}
  \caption{Sample images with corresponding ground truth feature correspondences from the WILLOW-ObjectClass data set.}
  \label{fig:X:2}
\end{figure*}

\noindent
\paragraph{WILLOW-ObjectClass data set}.
Next we follow the procedure in~\cite{qianGit} to perform multi-image matching for the WILLOW-ObjectClass data set~\cite{willowDataSet}. Sample images are shown in Fig.~\ref{fig:X:2}. We compare \textbf{Our} method in terms of F-score with \textbf{PG}~\cite{Wang:2017ub}, \textbf{Spectral}~\cite{Pachauri:2013wx}, \textbf{MatchLift}~\cite{Chen:2014uo}, \textbf{MatchALS}~\cite{zhou2015multi}, \textbf{NmfSync}~\cite{bernard2019synchronisation}, and \textbf{SparseStiefelOpt}~\cite{bernard2021sparse}. Except \textbf{Our}, all methods use the known ground-truth universe size for the multi-matching. For \textbf{Our} we use $k=100$, $\epsilon_{\alpha} = 0.01$, $N=20$, $\epsilon_{\eta} = 0$, $\kappa = 0$. Results are shown in Table~\ref{tab:table1}, where it can be seen that our method is competitive  without knowing the universe size. \\

\begin{table}
  \centering
  \setlength{\tabcolsep}{4.5pt}
  \begin{tabular}{@{}ccccccc@{}}
    \toprule
       & \textbf{Car} & \textbf{Duck} & \textbf{Face} & \textbf{Motorbike} & \textbf{Winebottle} & \textbf{universe size}\\
    \midrule
      Our &  0.985 & 0.935 & 1 & 0.983 & 1 & estimated\\
      PG  &  0.990 & 0.925 & 1 & 1 & 1 & ground truth\\
      Spectral & 0.990 & 0.929 & 1 & 0.975 & 1 & ground truth\\
      Matchlift & 0.990 & 0.925 & 1 &  0.976 & 1 & ground truth\\
      MatchALS & 0.980 & 0.921 & 1 & 0.960 & 1& ground truth \\
      NmfSync & 0.990 & 0.925 & 1 &  0.975 & 1 & ground truth\\
      SparseStiefelOpt & 0.980 & 0.861 & 1 & 0.975 & 1 & ground truth\\
    \bottomrule
  \end{tabular}
  \vspace{2mm}
  \caption{WILLOW dataset F-score comparison for different methods. Horizontally: image data sets. Vertically: methods. }
  \label{tab:table1}
\end{table}

\begin{figure*}[!h]
\vspace{-10pt}
  \includegraphics[width=\linewidth]{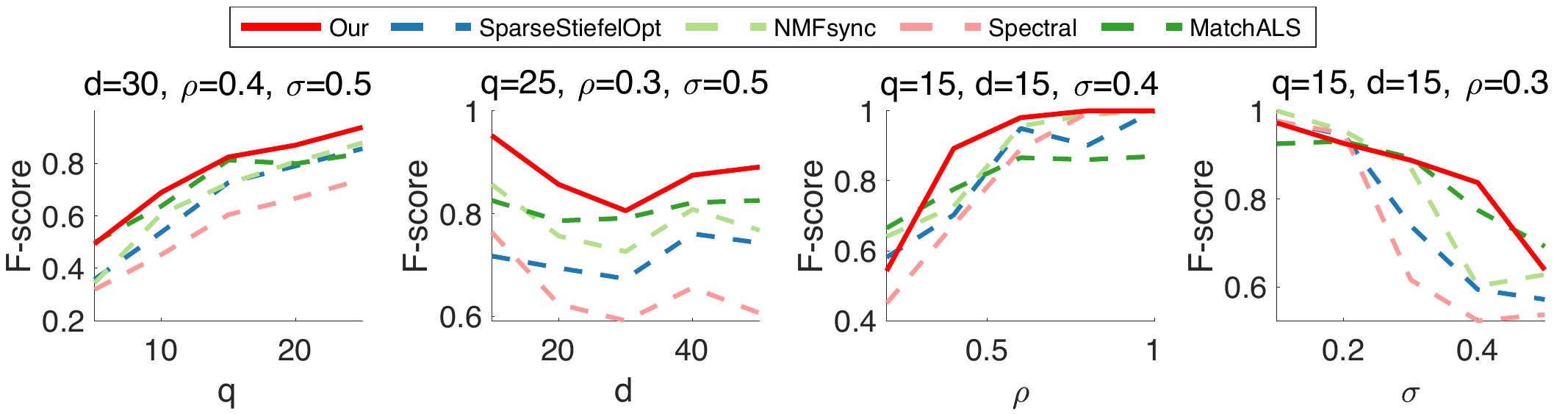}
  \vspace{-20pt}
  \caption{F-score ($\uparrow$) comparison of \textbf{Our} (Red) and four other methods for synchronization of partial permutation matrices. Methods with {dashed lines use ground-truth universe size} (i.e. $d$) in the matching, whereas the method with solid line (\textbf{Our}) does not. 
  In each setting three of the parameters $(q,d, \rho, \sigma)$ are constant, whereas the fourth is varying along the x-axis. The y-axis represents the average F-score over 50 problem instances for each choice of the varying parameter. All methods except \textbf{Our} use the universe size for the synchronization.}
  \label{fig:X:22}
\end{figure*}

\noindent
\paragraph{Multi-matching with partial observations}.
We continue with experiments on simulated (synthetic) problems where only a subset of features are observed  for each of the $q$ objects (corresponding to images). This means that the number of points $m_i$ is not necessarily equal to $m_j$ for $i \neq j$. We conduct a series of experiments where we generate problems using the following parameters. $q$: number of objects; $d$: ground truth universe size; $\rho$: probability that a feature is observed (for $\rho = 0$ it holds $m_i = 0$ for all $i$, and for $\rho =1$ it holds $m_i = d$ for all $i$); $\sigma$: this parameter specifies the degree of error in the similarity matrix $W$. It is the portion of randomly shuffled ground truth matchings for pairs of objects. An evaluation for different parameter settings is shown in Fig.~\ref{fig:X:22}, where average F-scores over 50 problems are presented for each parameter setting. 

\subsection{Clustering}
\noindent
\paragraph{Random binary similarity matrices.}\label{sec:RBS}
We begin the clustering evaluation by considering binary similarity matrices as input, which can be seen as the adjacency matrix of an unweighted graph, and hence the setting is in line with graph clustering. The similarity matrix represents matchings between $m$ points randomly assigned to one of the $k^*$ clusters.
In this context, two parameters capture the noise level of the similaity matrices considered:  $0 \leq \rho \leq 1$ and $0 \leq \nu \leq 1$, where the former captures the degree of randomness in the elements, and the latter captures the degree of missing data (i.e.~the degree of connectivity of the graph). Higher values are more challenging. 

\begin{figure*}[!h]
  \includegraphics[width=\linewidth]{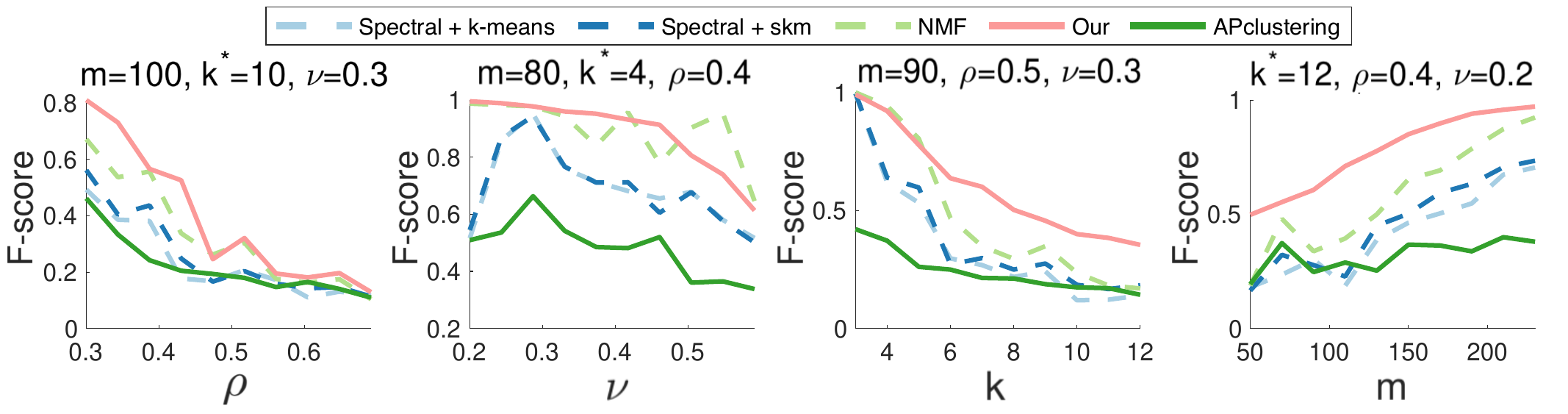}
  \vspace{-12pt}
  \caption{F-score ($\uparrow$) comparison of \textbf{Our} (pink) and five other clustering methods.
  In each setting three of the parameters $(m,k^*, \rho, \nu)$ are constant, whereas the fourth is varying along the x-axis. The y-axis represents the average F-score over 50 problem instances for each choice of the varying parameter. Methods with solid lines assume no knowledge of $k^*$, whereas methods with dashed lines use $k^*$ as the number of clusters to determine.}
  \label{fig:4}
\end{figure*}

We compare our method to existing clustering methods in this setting. Results are provided for four different settings in Fig.~\ref{fig:4}. In each of the four settings one parameter varies, whereas the others are fixed. The average F-scores are shown for 50 simulations for each choice of the varying parameter. 
The  algorithms we compare are the following ones. \textbf{Our}: Algorithm~\ref{alg:4} and Algorithm~\ref{alg:5}; \textbf{APclustering: } Affinity propagation clustering~\cite{bodenhofer2011apcluster,frey2007clustering}; \textbf{NMF}: Non-negative matrix factorization of $W$~\cite{Gaujoux1} with multiplicative updates~\cite{lee2001non} and simple rounding; \textbf{Spectral + kmeans}: Spectral clustering with k-means clustering~\cite{fcas} using rounding as in~\cite{ng2002spectral}; \textbf{Spectral + skm}: same as \textbf{Spectral + kmeans} except rounding with spherical k-means clustering~\cite{Kober1}. Of the compared methods, only \textbf{Our} and \textbf{APclustering} (solid lines) do not use $k^*$ explicitly in the clustering. Regarding \textbf{Our}, we have chosen the following parameters in all experiments in Fig.~\ref{fig:4}:  $k=100$, $\epsilon_{\alpha} = 0.01$, $N=20$ $\epsilon_{\eta} = 0$, $\kappa = 0$. \\

\noindent 
\paragraph{Gaussian mixture models.}\label{sec:GMM}
\begin{figure}[h]
\centering
  \includegraphics[scale=0.75]{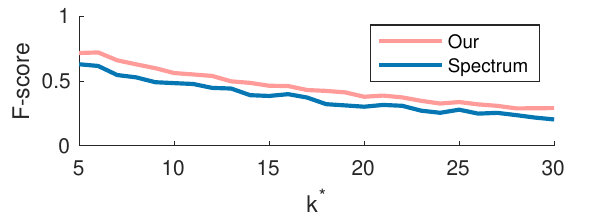}~~
  \vspace{-8pt}
  \caption{Quantitative comparison in terms of the mean F-Score ($\uparrow$) between our method and \textbf{Spectrum} for varying number of mixture components $k^*$.}
  \label{fig:gm1}
\end{figure}
Here we consider data from 2D Gaussian mixture models (GMM), where the number of mixture components is $k^*$ and the $m$ rows of a data matrix $X$ correspond to $m$ observations. The generated problems are challenging with highly overlapping clusters and different scales.
We used the adaptive kernel proposed by~\cite{john2020spectrum} to construct a similarity matrix $W$ and used their method \textbf{Spectrum} provided in~\cite{fcas,thrun2020clustering} as comparison method, which is a self-tuning spectral clustering method using the adaptive kernel, followed by a procedure for selecting the appropriate number of clusters, as well as a final step of Gaussian mixture model fitting.  Results are shown in Fig.~\ref{fig:gm1}, where we vary the number of $k^*$ ground truth clusters. For each choice of $k^*$, we randomly generate $30$ GMMs. In this generation of models, the mean distance between cluster centers is $1$ and mean variance within clusters is $0.7$. For each GMM we generate 200 sample points to create a problem for evaluation. 

\subsection{Discussion and Limitations}\label{sec:lim}
Once we know the index $i^*$ from Algorithm~\ref{alg:5}, we can run Algorithm~\ref{alg:4} again for $i^*$ many steps in the outer loop and then terminate. An alternative is to save the cluster assignments after each iteration in a vector.  For clustering problems, the rounding procedure does not comprise a computational bottleneck (its computational complexity is  $\mathcal{O}(km)$). Thus it can be added in each iteration of the outer loop to observe how the number clusters vary with varying $\alpha$.  
From a computational perspective the per-iteration complexity of Algorithm~\ref{alg:1} is equivalent to that of the orthogonal iteration algorithm for computing eigenvectors. The computational time is reasonable for moderate problem sizes. For example, the largest computational time observed for all problems used for Fig.~\ref{fig:4} in Section~\ref{sec:experiments}, was 1.34 seconds on a 2,7 GHz Intel Core i7 MacBook Pro.  
However, the proposed method suffers from some shortcomings from a computational perspective. Increasing $N$ and decreasing $\epsilon_{\alpha}$ results in more iterations (the total number being $N$ times the number of discretization steps of $\alpha$). The optimal initialization with $U_0 =K$ helps to reduce the $N$ needed to get a good solution. However, the eigenvalue computation in each iteration could provide a computational bottleneck. This can be alleviated by using more efficient approximation methods and additionally use a previously computed eigenvector as initialization for the next iteration in the outer loop.

\section{Conclusion}
In this work have presented a novel non-negative spherical relaxation for multi-matching and clustering without the knowledge of universe size. An efficient and easy-to-implement conditional power iterations method is used for optimization, where a continuous parameter is selected by comparing convergence rates for values thereof. The solution for this parameter-choice is subsequently rounded to determine the assignments. The method is competitive against state-of-the art methods that assume  knowledge of the universe size.

\subsubsection*{Acknowledgements.} The authors gratefully acknowledge the financial support from the Swedish Research Council (2019-04769).

{\small
\bibliographystyle{splncs04}
\bibliography{paper}

\newpage

\appendix

\section{Proof of Proposition~1}
\noindent
1. It holds, by definition of $U_{t+1}$ that \textbf{a)} 
\begin{equation}
\label{eq:17}
 \text{tr}(U_t^T V(\alpha) U_{t+1}) \geq \text{tr}(U_t^T V(\alpha) U_{t}).
\end{equation}
Moreover, since $V(\alpha)$ is positive definite, it holds that \textbf{b)}
\begin{align}
\label{eq:18}
0 & \leq \text{tr}((U_{t+1} - U_{t})^TV(\alpha)(U_{t+1} - U_{t})) \\
& =  \text{tr}(U_{t+1}^T V(\alpha) U_{t+1}) +  \text{tr}(U_{t}^T V(\alpha) U_{t}) - 2\text{tr}(U_{t}^TV(\alpha)U_{t+1}).
\end{align}
Combining  \textbf{a)} and \textbf{b)} yields the desired inequality. \\

\noindent 
2. Since $\mathcal{S}$ is compact and $f(U_t) = \text{tr}(U_{t}^TV(\alpha)U_{t})$ is increasing (see 1. above), $f(U_t)$ converges to a point as $t$ goes to $\infty$. \\

\noindent 
3. For 
symmetric and positive definite
matrices in $\mathbb{R}^{m \times m}$ we define the inner product as $\langle X, Y \rangle_{{V(\alpha)}} = \text{tr}(X^TV(\alpha)Y)$, 
and the norm $\|\cdot\|_{V(\alpha)} = \sqrt{\text{tr}((\cdot)^TV(\alpha)(\cdot)}$. We prove the statement with this alternative norm, whereby it  holds for the  Frobenius norm due to the equivalence of the two norms.  

We know from 1) and 2) that $\|U_t\|^2_{V(\alpha)}$ monotonically converges to a $\delta > 0$ from below, as $t$ goes to infinity. Thus, for $\epsilon > 0$ there is a $T$ such that $\delta - \|U_t\|^2_{V(\alpha)} \leq \epsilon$ for $t \geq T$. We note that, as a consequence of the relations \eqref{eq:17} and \eqref{eq:18}, the following is true (for all $t \geq 0$):
\begin{equation}\label{eq:19}
\|U_{t+1} - U_{t}\|^2_{{{V(\alpha)}}} \leq \|U_{t+1}\|^2_{{{V(\alpha)}}} - \|U_{t}\|^2_{{{V(\alpha)}}}.
\end{equation}
For $t > T \geq 0$ we define the partial sum $s_{T,t}$ as below. On the right-hand side, we use \eqref{eq:19} (repeatedly) to obtain the  inequality.
\begin{equation}
s_{T,t} := \sum_{l= T}^{t-1}\|U_{l+1} - U_l\|^2_{V(\alpha)} \leq \|U_{t}\|^2_{{{V(\alpha)}}} - \|U_{T}\|^2_{{{V(\alpha)}}} \leq \epsilon.
\end{equation}
Now, since $s_{T,t}$ is monotonically increasing 
for increasing $t$ and bounded from above by $\epsilon$, there is $s^* \leq \epsilon$ s.t. $\lim_{t \rightarrow \infty}s_{T,t} = s^*$. \hfill$\square$
}

\end{document}